 \documentclass[pmlr,twocolumn,10pt]{jmlr} 

\usepackage{listings}
\usepackage{booktabs}
\usepackage{siunitx}
\usepackage{threeparttable} 
\usepackage{tcolorbox}

\usepackage{subcaption} 
\usepackage{floatrow}  
\usepackage{makecell} 
\usepackage{hyperref}
\usepackage{lineno}
\newcommand{\add}[1]{\textcolor{black}{#1}}

\newcommand{\equal}[1]{{\hypersetup{linkcolor=black}\thanks{#1}}}

\theorembodyfont{\upshape}
\theoremheaderfont{\scshape}
\theorempostheader{:}
\theoremsep{\newline}

\jmlrvolume{297}
\jmlryear{2025}
\jmlrworkshop{Machine Learning for Health (ML4H) 2025} 
 \title[CancerGUIDE]{CancerGUIDE: Cancer Guideline Understanding via Internal Disagreement Estimation}

\author{%
\Name{Alyssa Unell} \Email{aunell@stanford.edu}\\
\addr Microsoft Research \and Stanford University
\AND
\Name{Noel C. F. Codella}\equal{Project Mentors} \Email{ncodella@microsoft.com}\\
\addr Microsoft Health and Life Sciences
\AND
\Name{Sam Preston}\footnotemark[1] \Email{sam.preston@microsoft.com}\\
\addr Microsoft Health and Life Sciences
\AND
\Name{Peniel Argaw}\footnotemark[1] \Email{penielargaw@microsoft.com}\\
\addr Microsoft Research
\AND
\Name{Wen-wai Yim} \Email{yimwenwai@microsoft.com}\\
\addr Microsoft Health and Life Sciences
\AND
\Name{Zelalem Gero} \Email{zelalemgero@microsoft.com}\\
\addr Microsoft Research
\AND
\Name{Cliff Wong} \Email{cliff.wong@microsoft.com}\\
\addr Microsoft Research
\AND
\Name{Rajesh Jena} \Email{t-rajeshjena@microsoft.com}\\
\addr Microsoft Research
\AND
\Name{Eric Horvitz} \Email{horvitz@microsoft.com}\\
\addr Microsoft
\AND
\Name{Amanda K. Hall} \Email{amanda.hall@microsoft.com}\\
\addr Microsoft Research
\AND
\Name{Ruican Rachel Zhong} \Email{t-rzhong@microsoft.com}\\
\addr Microsoft Research \and University of Washington
\AND
\Name{Jiachen Li} \Email{t-jiachenli@microsoft.com}\\
\addr Microsoft Research
\AND
\Name{Shrey Jain} \Email{shreyjain@microsoft.com}\\
\addr Microsoft Health and Life Sciences
\AND
\Name{Mu Wei} \Email{muhsin.wei@microsoft.com}\\
\addr Microsoft Health and Life Sciences
\AND
\Name{Matthew Lungren}\equal{Equal Leadership} \Email{mlungren@microsoft.com}\\
\addr Microsoft Health and Life Sciences
\AND
\Name{Hoifung Poon}\footnotemark[2] \Email{hoifung@microsoft.com}\\
\addr Microsoft Research
}

\begin{document}

\maketitle

\begin{abstract}
The National Comprehensive Cancer Network (NCCN) provides evidence-based guidelines for cancer treatment. Translating complex patient presentations into guideline-compliant treatment recommendations is time-intensive, requires specialized expertise, and is prone to error. Advances in large language model (LLM) capabilities promise to reduce the time required to generate treatment recommendations and improve accuracy. We present an LLM agent-based approach to automatically generate guideline-concordant treatment trajectories for patients with non-small cell lung cancer (NSCLC). Our contributions are threefold. First, we construct a novel longitudinal dataset of 121 cases of NSCLC patients that includes clinical encounters, diagnostic results, and medical histories, each expertly annotated with the corresponding NCCN guideline trajectories by board-certified oncologists. Second, we demonstrate that existing LLMs possess domain-specific knowledge that enables high-quality proxy benchmark generation for both model development and evaluation, achieving strong correlation (Spearman coefficient $r=0.88$, RMSE = $0.08$) with expert-annotated benchmarks. Third, we develop a hybrid approach combining expensive human annotations with model consistency information to create both the agent framework that predicts the relevant guidelines for a patient, as well as a meta-classifier that verifies prediction accuracy with calibrated confidence scores for treatment recommendations (AUROC=0.804). Calibrated confidence scoring is a critical capability for communicating the accuracy of outputs, custom-tailoring tradeoffs in performance, and supporting regulatory compliance. This work establishes a framework for clinically viable LLM-based guideline adherence systems that balance accuracy, interpretability, and regulatory requirements while reducing annotation costs, providing a scalable pathway toward automated clinical decision support. 
\end{abstract}


\paragraph*{Data and Code Availability} We obtain clinical annotations from a third-party contractor. Patient data is de-identified and sourced from multiple institutions within the USA, but it is not licensed for public release. All analyses were conducted using Azure-hosted LLMs within our institution’s secure cloud environment; no data were transmitted to external third-party services. Code and synthetic patient data are made available here: \href{https://aka.ms/CancerGUIDE}{CancerGUIDE repository}.

\paragraph*{Institutional Review Board (IRB)}
Our research does not contain data from identified human subjects and as such does not require an IRB.

\section{Introduction}
Cancer treatment decisions require oncologists to synthesize complex patient histories with evolving clinical guidelines to recommend the appropriate next treatments. The National Comprehensive Cancer Network (NCCN) guidelines provide evidence- and consensus-based recommendations for cancer diagnosis, treatment, and management \citep{NCCN}, with adherence promoting higher quality and consistent care between providers \citep{Pluchino2020,Winn2003,Benson2008}. However, guideline navigation presents significant challenges: the guidelines are extensive, frequently updated as new research emerges, and require time-intensive review of complex patient documentation \citep{Benson2008}. These factors contribute to variability in guideline adherence and treatment recommendations, particularly in resource-constrained settings where specialist expertise is limited \citep{Koh2020,Kumar2023}. 

Large language models (LLMs) offer promising potential to address these challenges by automatically processing clinical notes and recommending guideline-concordant treatment plans \citep{Fast2024,Wang2024,Hamed2023,Schulte2023}. However, deploying LLMs for clinical decision support requires rigorous evaluation to ensure accuracy and safety. The complexity of clinical reasoning, combined with the high stakes of treatment decisions, demands robust validation methods that can assess model performance at scale while maintaining clinical safety standards. In addition, current FDA guidelines for evaluation of AI systems recommend ROC curve measurement as a part of comprehensive clinical performance assessment endpoints \citep{fda}, which is challenging to produce from generative outputs that are not associated with semantically aligned confidence scores. 

Despite growing interest in clinical LLMs, rigorous evaluation remains a fundamental bottleneck due to the scarcity of expert-annotated datasets. High-quality ground truth labels for complex clinical reasoning tasks require substantial investment in specialist time and expertise, limiting the scale at which models can be validated \citep{Zac2023,Liang2017}. Common evaluation approaches face significant limitations: synthetic data generation often fails to capture clinical complexity and is vulnerable to distributional shift \citep{Santangelo2025,Yan2022,Chen2024,TIMER}, while using actual patient treatments as ground truth is problematic, since real-world decisions frequently incorporate factors exogenous to guideline recommendations, such as patient preferences, drug availability, institutional protocols, and physician experience \citep{Arts2016,Quaglini2004}.

This evaluation challenge is particularly acute for guideline adherence tasks, where the gold standard requires expert oncologists to determine whether complex, multi-step clinical reasoning aligns with evidence-based recommendations. The resulting annotation bottleneck creates a critical gap: while LLMs show promise for clinical decision support, practitioners lack scalable methods to assess model reliability before deployment in high-stakes healthcare settings.
\begin{figure}[h]
    \centering    
    \includegraphics[width=.8\textwidth]{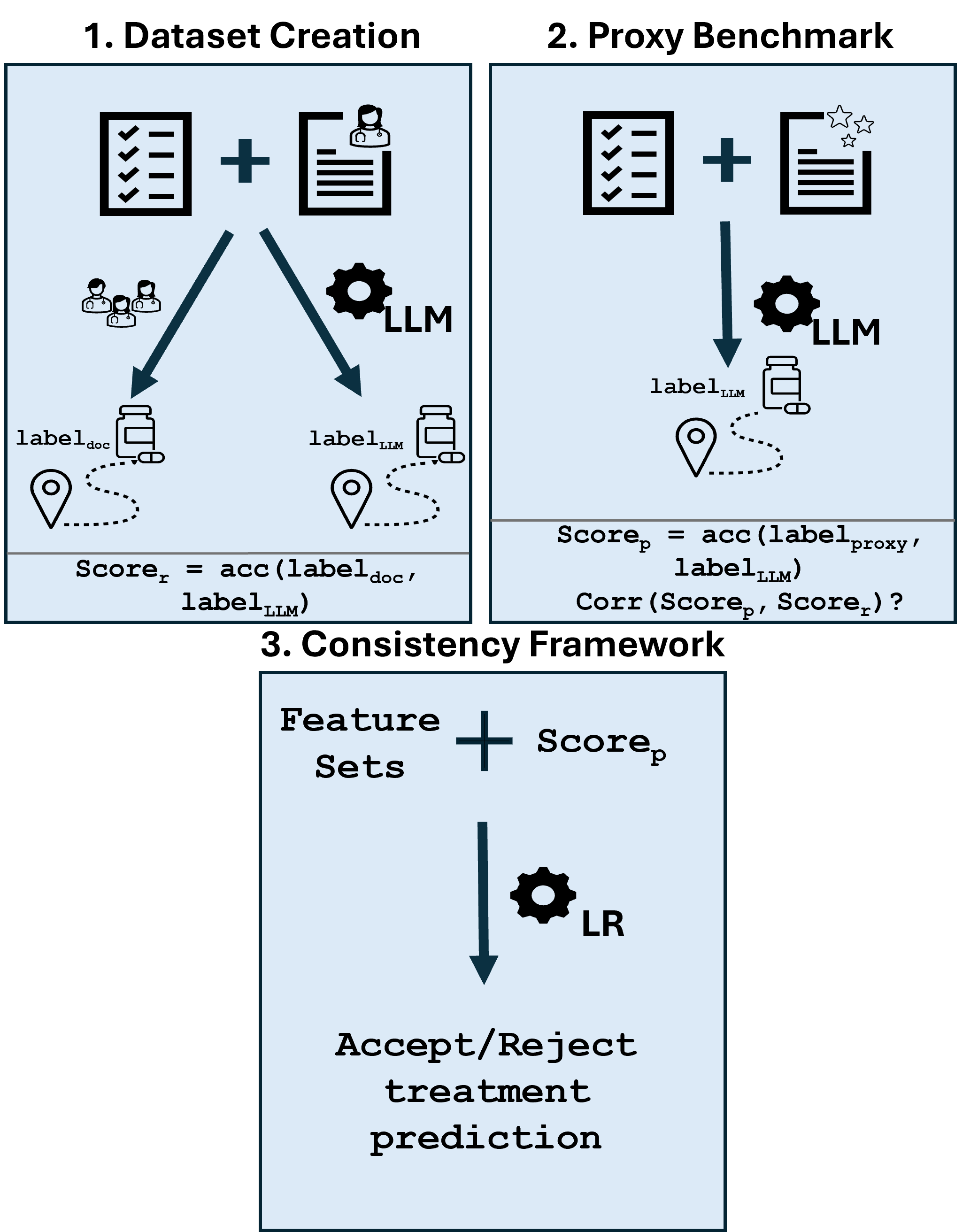} 
 \caption{
   \add{\textbf{CancerGUIDE framework.} (1) Clinicians derive patient pathways ($label_{doc}$) from NCCN guidelines and real notes. These serve as gold-standard references to compare with LLM-derived pathways ($label_{LLM}$), producing a reference accuracy score ($Score_r$). (2) NCCN guidelines and both synthetic and real clinical notes are used to generate weak labels ($label_{proxy}$). LLM predictions ($label_{LLM}$) are compared to these proxy labels to compute proxy performance scores ($Score_p$), enabling evaluation of how well synthetic supervision approximates expert annotations.   (3) Model consistency features (Feature Sets) and $Score_p$ are used to train a logistic regression meta-classifier that predicts whether a treatment recommendation is likely correct. This classifier is fit on labelled data from step 1 and unlabelled features from step 2. The classifier is applied at test time to accept or reject LLM-generated recommendations, supporting confidence estimation and threshold selection for clinical deployment.
}}
    \label{fig:overview}
\end{figure}

We address this evaluation bottleneck through two complementary approaches that enable scalable assessment of LLM performance on guideline adherence without extensive expert annotation. First, we evaluate models across six different proxy-benchmark generation methods: two synthetically generated datasets and four based on real clinical notes with consistency-derived labels. This analysis enables model selection and preliminary capability assessment without ground-truth labels. Second, we demonstrate that self- and cross-model consistency (the degree to which a model agrees with itself and other models) serve as reliable predictors of accuracy on expert-annotated cases.

To validate these approaches, we construct the first benchmark for NCCN guideline adherence on non-small cell lung cancer (NSCLC) by eliciting 13 oncologists to annotate 121 complete patient pathways, representing 130+ hours of specialist expertise. This expert-validated dataset enables us to systematically evaluate six different proxy benchmarking methods and quantify the relationship between model consistency and clinical accuracy across multiple frontier LLMs. We then develop a meta-classifier framework that combines these weak supervision signals to classify individual prediction correctness, achieving robust performance while requiring minimal expert validation (Figure \ref{fig:overview}).

Our evaluation highlights the role of synthetic data and consistency-based data in proxy benchmarking, with these approaches achieving high Spearman correlation coefficients (r=0.88) and low RMSE values (RMSE=0.08). We demonstrate that model consistency can also serve as a reliable accuracy predictor, enabling our meta-learning framework to achieve an average of 0.804 AUROC in classifying individual prediction accuracy across all models, while unsupervised clustering using consistency signals alone achieves 0.666 F1 at separating correct from incorrect predictions.

Our primary contributions are as follows:
\begin{enumerate}
    \item \textbf{NCCN guideline-adherent dataset and task formalization}: The first rigorous ML problem formulation for clinical guideline adherence, with an expert-annotated dataset of 121 patient pathways and benchmarking across eight frontier LLMs.
    \item \textbf{Proxy benchmark to validate performance in zero-label settings}: A systematic evaluation that identifies which proxy methods best predict clinical performance without expert labels.
    \item \textbf{Consistency framework for reliable treatment prediction}: A hybrid agent and meta-classifier framework that generates relevant guideline paths and produces confidence values correlated with accuracy, enabling ROC curve calculation to clearly convey performance to clinicians and ensure compliance with regulatory standards \citep{fda}.
\end{enumerate}

\section{Related Work}
\subsection{Guideline Adherent Treatment Recommendations}
Guideline-adherent treatments have consistently been associated with improved clinical outcomes, such as overall survival \citep{Dholakia2020,Lindqvist2022,Wckel2010}. Clinical decision-support systems have the ability to translate clinical guidelines into point-of-care recommendations. Several studies have demonstrated the use of AI-driven systems in predicting guideline-adherent treatment recommendations in oncology \citep{Nafees2023,Suwanvecho2021}. Emerging work has also explored the utility of LLMs; for example, preliminary investigations suggest that ChatGPT can effectively summarize guideline content \citep{Hamed2023} and exhibits partial concordance in identifying guideline-adherent treatments \citep{Schulte2023}. Furthermore, recent benchmarking efforts have systematically evaluated the alignment of LLMs with established medical guidelines, underscoring both their potential utility and the current limitations imposed by the relatively small number of verified clinical cases available \citep{Fast2024}.

\subsection{Synthetic Generation of Clinical Data}
Synthetic data generation is a promising strategy to address the challenges of privacy and data scarcity in healthcare. Past works have focused on generating high-fidelity electronic health records (EHRs) using generative models, such as generative adversarial networks, variational autoencoders and autoregressive models \citep{Choi2017,Lu2022,Li2023,Yoon2023,Jarret2021}. More recent frameworks have used LLMs to generate synthetic data and have shown improvements in privacy guarantees and scalability \citep{Altalla2025,Barr2025,Taloni2025,Litake2024}. In parallel, standardized evaluation metrics and scorecards have been proposed to assess the fidelity, privacy, and clinical utility of synthetic health data, providing a foundation for more rigorous benchmarking and deployment \citep{Santangelo2025,Yan2022,Chen2024, medhelm}. In addition, synthetic cohorts have been developed to model guideline-adherent treatment pathways, such as synthetic stroke registries for adherence benchmarking \citep{Wang2025} and synthetic EHR modules that embed clinical practice guidelines and protocols \citep{Walonoski2017}. \add{Previous work has explored guideline-following capability across frontier models on synthetic clinical data }\citep{li2025medguide}. These efforts demonstrate the potential of synthetic data for adherence-focused tasks. However, significant challenges remain in generating high-fidelity, guideline-adherent cases while maintaining performance on downstream evaluation tasks.

\subsection{Weak Supervision and Consistency}
Weak supervision and consistency-driven learning have emerged as key strategies for training models when labeled data is limited or noisy. Weak supervision techniques, such as programmatic labeling and distant supervision, integrate heterogeneous, partially labeled sources to produce probabilistic labels for downstream predictive tasks \citep{Mintz2009,Ratner2016,Ratner2017,Hsu2025,Falcon2025}. Consistency-based methods encourage models to produce stable predictions under input perturbations, data augmentations, or repeated evaluations, helping to regularize training and improve generalization \citep{Laine2016,Miyato2019,Xie2020,monkeys}. \add{Cycle consistency is a related method for generating weakly supervised labels from one data source. It involves transforming data into labels and then mapping those labels back to the original data to measure how well the label preserves the original information, providing a confidence estimate for the weakly supervised label \citep{cycleconsistency}}.
\emph{Pseudo-labeling}, a related strategy, generates labels for unlabeled or partially labeled data by treating confident model predictions as additional supervised training data, enhancing model performance by expanding the training set with its own high-confidence predictions \citep{Lee2013,Zou2021,Xie2020}. Within healthcare, pseudo-labeling has shown to improve the reliability of EHR phenotyping and imaging tasks under label scarcity \citep{Dong2023,Zhang2022,Nogues2022}. Together, these techniques provide a framework for leveraging limited labeled data while maintaining prediction stability and adherence to domain-specific constraints, which is particularly relevant for guideline-adherent treatment modeling.

\subsection{Non-Verifiable Task Evaluation}
Evaluating LLMs on open-ended or non-verifiable tasks, such as treatment recommendation prediction, is challenging due to the scarcity of ground truth labels. Traditional metrics like accuracy or BLEU are insufficient in zero-label settings, prompting the use of human-aligned and/or proxy evaluation frameworks \citep{Lyu2024,Yamauchi2025,Tan2024}. Further, studies have shown that consistency-derived benchmarks provide improved predictive fidelity relative to ground-truth outcomes \citep{Lee2024,Li2025}. Proxy frameworks and consistency-derived benchmarks offer partial solutions, but their alignment with expert judgment remains imperfect, motivating the continued development of hybrid evaluation strategies.

\section{Methods}
\subsection{Guideline-Compliance Task Formalization}
We formalize guideline-compliant treatment prediction as a structured prediction problem. Let $x \in \mathcal{X}$ denote patient notes and $y \in \mathcal{Y}$ the corresponding guideline-compliant pathway, where $\mathcal{Y}$ is the space of decision graphs with nodes as clinical decisions and terminal nodes as treatments. An LLM $f: \mathcal{X} \to \mathcal{Y}$ produces predictions $\hat{y} = f(x)$.  

To ground this formalization in clinical practice, we leveraged o3’s vision capabilities \citep{openai2025o3} to extract the NCCN NSCLC guideline decision tree \citep{NCCN} and curated an expert-annotated dataset. Oncologists traced patient notes through the decision tree, recording node sequences and assessing guideline adherence for each case. Further details on annotation procedures, participant recruitment, and quality control measures are provided in Appendix~\ref{subsec:annotation}.

Since gold-standard pairs $(x,y)$ are difficult to obtain, direct supervision is not scalable as $X$ grows. To address this, we first introduce \emph{proxy benchmarking} methods that act as substitutes for direct supervision. We define synthetic inputs $\hat{x}$ and corresponding predicted pathways $\hat{y}$, forming two primary classes of proxy evaluations: 
\begin{enumerate}
    \item \textbf{Proxy using synthetic inputs:} $(\hat{x}, \hat{y})$, where $\hat{x}$ represents synthetic patient notes generated conditionally on guideline paths. This allows assessment of model reliability under controlled perturbations or alternative representations of patient data.  
    \item \textbf{Proxy using real inputs:} $(x, \hat{y})$, where the model prediction $\hat{y}$ is generated multiple times by $f$ and the  $(x, \hat{y})$ pair is kept only if minimum self-consistency is reached.  
\end{enumerate}
These proxy-based evaluations provide measurable signals of model performance, enabling ranking of models or detection of likely errors in zero-label settings. We learn a surrogate evaluator $g$ that predicts whether $\hat{y}$ is guideline-compliant. Concretely, we define a feature mapping
\[
\phi: (\mathcal{X}, \mathcal{Y}, f) \to \mathbb{R}^d
\]
that extracts signals such as (i) model self-consistency across rollouts, (ii) agreement across models, and (iii) alignment with proxy benchmarks. The evaluator is then trained as a binary classifier
\[
g(\phi(x,\hat{y},f)) \approx \mathbf{1}\{\hat{y} = y\}.
\]

Our objective is to minimize the expected classification loss
\[
\min_{g} \; \mathbb{E}_{(x,y) \sim \mathcal{D}} \big[ \mathcal{L}(g(\phi(x,\hat{y},f)), \mathbf{1}\{\hat{y}=y\}) \big],
\]
where $\mathcal{D}$ denotes the distribution over patients and $\mathcal{L}$ is standard loss. This formulation enables evaluation of $f$ in settings with limited or no access to human-labeled $(x,y)$ pairs, by leveraging model agreement, self-consistency, and benchmark proxy-derived features as signals for meta-classification.

\subsubsection{Evaluation Preliminaries}
To evaluate model performance, we employ two complementary metrics:
\begin{enumerate}
\item \textbf{Path Overlap}: Measures the proportion of nodes in predicted paths that are repeated, relative to the total nodes in all compared paths. This captures consistency of decision sequences with the full derivation in Appendix~\ref{subsec:path_overlap}
\item \textbf{Treatment Match}: A binary score indicating whether the final predicted treatment matches the ground truth when available. In settings without ground truth, it is computed as the proportion of repeated final treatments across multiple predictions, with full derivation in Appendix~\ref{subsec:treatment_match}.
\end{enumerate}
\add{Path Overlap captures the model's ability to navigate the guideline-adherent decision process even if the final treatment is not fully correct, reflecting potential utility and acceptability in assisting clinicians. Treatment Match provides a direct measure of clinical accuracy for the final recommendation. Together, these metrics offer complementary perspectives: one emphasizes process fidelity and decision support potential, while the other captures correctness of the outcome. To ensure consistent and deterministic scores for path quality, LLM judges are not used to assess similarity of paths as this introduces stochasticity between models that confounds direct comparison in this setting.}

\subsection{Zero-Label Benchmark Generation}
\label{subsec:zero-label}
To approximate $(x,y)$ pairs for evaluation, we introduce two complementary approaches (\textbf{synthetic supervision} and \textbf{consistency-based pseudo-labeling}), with a total of six proxy benchmarking methods for zero-label performance estimation contributed.

\subsubsection{Synthetic Supervision}
We generate high-fidelity synthetic patient notes $\hat{x}$ paired with generated guideline paths $\hat{y}$ to simulate real clinical cases. The goal is to maximize fidelity to realistic patient notes while ensuring the target guideline path is accurate. Our multi-step pipeline separates the generation of $\hat{x}$ and the selection of $\hat{y}$ to filter incorrect labels from the benchmark, while still maintaining meaningfully realistic patient cases.

\subparagraph{Generating $\hat{x}$} 
Two complementary strategies are used: \textbf{Structured Generation} fills empty structured fields conditioned on the generated path and full clinical guidelines, performs consistency checks by reconstructing implied paths (discarding mismatched cases), then generates unstructured notes from structured data, target path, guidelines, and real clinical note examples. \textbf{Unstructured Generation} bypasses structured fields, generating synthetic notes directly from target paths, guidelines, and clinical note examples. \add{Synthetic datasets were generated with GPT-4.1, except for those used to evaluate GPT-4.1, which were produced by GPT-5 under minimal-reasoning conditions.} 

\subparagraph{Selecting $\hat{y}$ via LLM Preference}
Once $\hat{x}$ is generated, we obtain $\hat{y}$ by having the LLM generate the path from $\hat{x}$. If the prediction matches the target path, we accept the pair directly. Otherwise, the LLM chooses between predicted and target paths in position-agnostic format, and we retain only cases where the target path is selected.

The final dataset is then composed of $(\hat{x}, \hat{y})$ pairs which the generation model either correctly regenerated $\hat{y}$ from $\hat{x}$ or was able to select $\hat{y}$ from a pair of available $(\hat{y}^*, \hat{y})$, with $\hat{y}^*$ being the path prediction generated for $\hat{x}$. This captures examples where direct generation fails but verification remains feasible (e.g., generating the correct path is difficult, but identifying it is easier) \citep{verif1, Falcon2025}. 

\subsubsection{Consistency-Based Pseudo-Labeling}
We propose a consistency-based pseudo-labeling strategy to construct proxy benchmarks that approximate model performance on treatment prediction and guideline path generation. Consistency-based pseudo-labels are derived from two sources: (i) \textit{Self-consistency}, which leverages agreement within repeated predictions of a single model, and (ii) \textit{Cross-Model Consistency}, which relies on agreement across different models. For each source, we define two benchmarks based on whether consistency is measured with respect to the \textit{path overlap} metric or the \textit{treatment match} metric, yielding four benchmarks in total. 

\add{\paragraph{Self-Consistency}
For real clinical notes, pseudo-labels are generated using model self-consistency. Specifically, we sample $k$ independent predictions $f(x) = {\hat{y}_1, \dots, \hat{y}_k}$ from the same model $m$ across $X$ questions. Agreement among these predictions is assessed along two axes: \textit{path overlap} (structural alignment) and \textit{treatment match} (final treatment concordance). Notes with agreement above a threshold $\delta = 0.9$ on the target metric are retained, while inconsistent cases are assigned a score of $0$. The retained $(x, \hat{y})$ pairs are then used to evaluate model $m$, where $\hat{y}$ corresponds to the most frequent prediction among $f(x)$. Performance is computed as accuracy over the retained subset of $X$, counting inconsistent answers as incorrect.}

\paragraph{Cross-Model Consistency}
We further assign pseudo-labels by comparing predictions across  the whole set of models, $M$. For a note $x$, if two or more models converge on the same pseudo-label after $k$ independent samples, we define this as the aggregated label. \add{Convergence in this setting is exact path match.} Inconsistent cases are simply excluded rather than penalized in contrast to the Self-Consistency method. \add{We obtain the subset of (x, $\hat{y}$) pairs with the aggregated labels and evaluate all models in $M$ on this proxy benchmark}.

\subsection{Final Treatment Accuracy Prediction}
\paragraph{Supervised Accuracy Classification} \add{We use consistency signals and model performance on proxy benchmarks to predict accuracy of a generated final treatment prediction}. We train a meta-classifier using features derived from self-consistency metrics ($k$-rollout path overlap and treatment match), cross-model consistency (fraction of models with the same generated path), and proxy benchmark scores (synthetic and consistency-derived). 

\paragraph{Unsupervised Performance for Accuracy Classification and Error Identification} To demonstrate label-free evaluation capabilities, we apply unsupervised methods to both accuracy classification and error identification. Clustering self- and cross-model consistency features naturally separates high-confidence correct from low-confidence incorrect predictions. 
\add{For error identification, we tabulate inconsistencies across $k$ rollouts for each model $m$ on patient $p$. This approach leverages consistency signals to detect potential errors without relying on human labels.}

\section{Results}
\subsection{Expert-Annotated Dataset}
To evaluate LLM performance on guideline-concordant treatment prediction, we constructed a high-quality expert-annotated dataset. 

\paragraph{Annotation Details}
A total of 13 oncologists annotated 121 patient notes, \add{averaging 46.5 minutes per patient note with a cost of \$500 per US Board certified clinician hour}. To assess inter-annotator reliability, 11 examples were dually annotated, showing an average of 0.636 treatment match and 0.692 path overlap score, \add{with further disagreement analysis present in Appendix~\ref{sec:disagreement_analysis}}. Notes were on average 54755 characters long and contained 82 distinct paths through the NCCN decision tree and 48 distinct final treatment recommendations. We use this annotation as ground truth for both evaluating model performance and validating proxy benchmarking approaches. By establishing a high-quality reference standard, we can interpret LLM performance, detect systematic errors, and calibrate models in high-stakes medical settings without relying exclusively on expensive ongoing human annotation.


\paragraph{Model Performance on Human-Annotated Benchmark}
Table \ref{tab:true_performance} illustrates the results of eight frontier models on the expert-annotated dataset. We report performance on GPT-5 \citep{openai2025gpt5}, GPT-4.1 \citep{openai2025gpt41}, o3 \citep{openai2025o3}, o4-mini \citep{openai2025o4mini}, DeepSeek-R1 \citep{deepseek2025r1}, and LLaMA-3.3-70B-Instruct \citep{meta2024llama3}. We evaluate GPT-5 with varying reasoning efforts (minimal, medium, and high) to directly assess the impact of reasoning on performance. All models are evaluated with default temperature 1.0. We measured two clinically relevant metrics: (i) \textit{path overlap}, measuring structural agreement with the annotated guideline path, and (ii) \textit{treatment match}, indicating whether the recommended treatment node matches expert annotation.

\begin{table}[t]
\centering
\caption{Model performance on Expert-Annotated Data (mean $\pm$ SEM).}
\scriptsize
\setlength{\tabcolsep}{4pt} 
\begin{tabular}{lcc}
\hline
\textbf{Model} & \textbf{Path Overlap} & \textbf{Treatment Match} \\
\hline
GPT-5-High & $0.455 \pm 0.035$ & $0.339 \pm 0.043$ \\
GPT-5-Medium & $0.483 \pm 0.036$ & $0.364 \pm 0.044$ \\
GPT-5-Minimal & $0.441 \pm 0.032$ & $0.322 \pm 0.043$ \\
GPT-4.1 & $0.388 \pm 0.035$ & $0.298 \pm 0.042$ \\
o3 & $0.477 \pm 0.038$ & $0.364 \pm 0.044$ \\
o4-mini & $0.433 \pm 0.037$ & $0.339 \pm 0.043$ \\
Deepseek-R1 & $0.419 \pm 0.038$ & $0.355 \pm 0.044$ \\
LLaMA-3.3-70B-Instr. & $0.174 \pm 0.022$ & $0.112 \pm 0.029$ \\
\hline
\end{tabular}
\label{tab:true_performance}
\end{table}

\subsection{Proxy Benchmarks Enable Model Evaluation in Zero-Label Settings}

\begin{figure}[htbp]
    \centering
    \includegraphics[width=\textwidth]{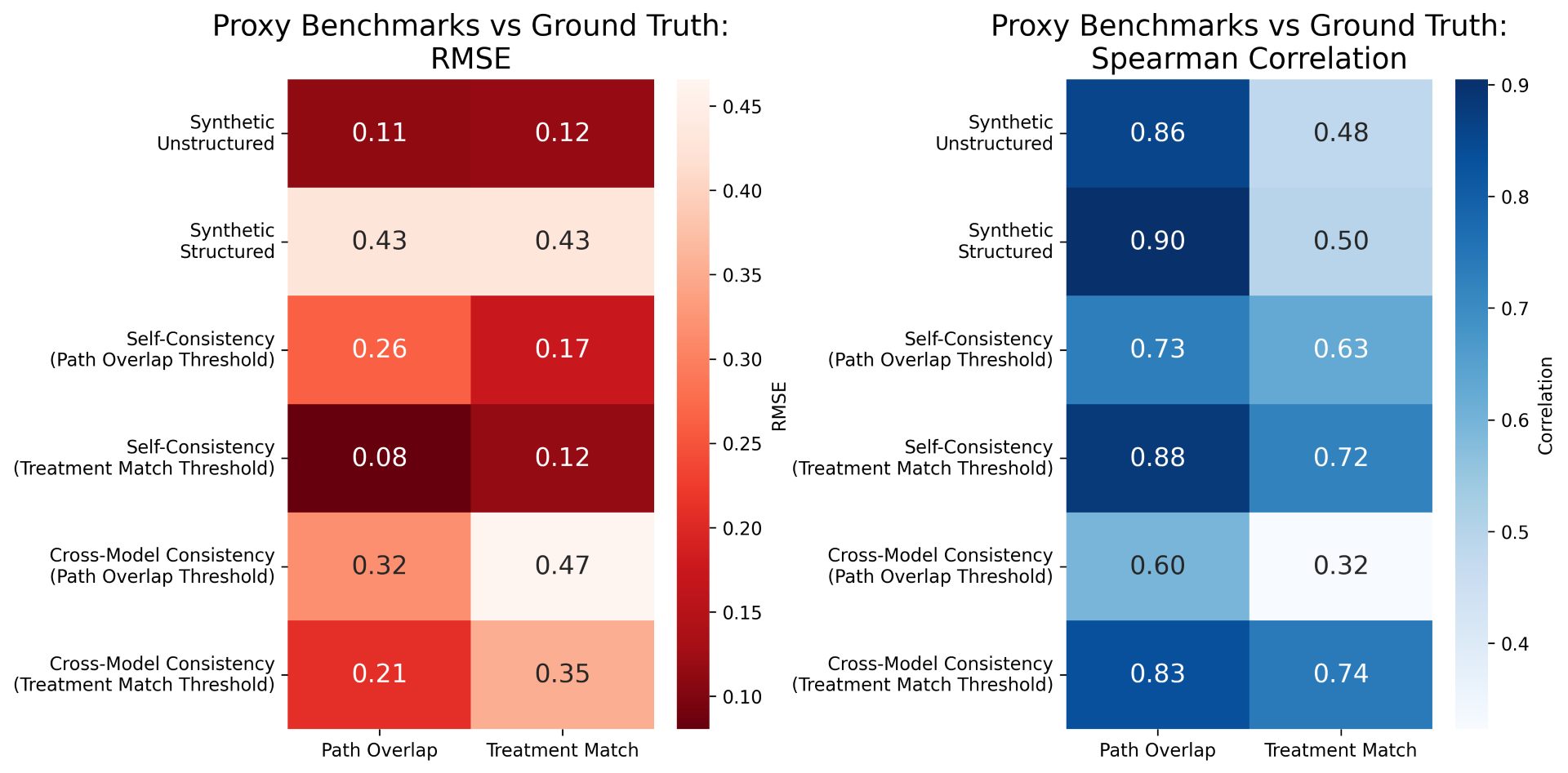} 
    \caption{\textbf{Self-consistency pseudo-labels provide a robust proxy for benchmarking.} Six approaches are evaluated: synthetic data (structured and unstructured), self-consistency pseudo-labeling (with varying acceptance criteria), and cross-model consistency pseudo-labeling. Correlation is measured using Spearman coefficients as well as root mean-squared error with color intensity indicating magnitude.}
    \label{fig:heatmap}
\end{figure}

We evaluated six proxy benchmarks for assessing LLM performance without human labels (Figure~\ref{fig:heatmap}). Among synthetic approaches, adding structure to the generation pipeline substantially increased error (RMSE rising from 0.11 in Synthetic Unstructured to 0.43 in Synthetic Structured) while yielding only a marginal correlation gain (Spearman $r$: 0.86 → 0.90). For consistency-based methods, thresholding by treatment match outperformed path overlap, likely because treatment match reflects a more generalizable prediction goal, whereas path overlap penalizes minor deviations that have little effect on downstream treatment accuracy. Aggregating outputs across models in cross-model consistency failed to improve correlations and instead increased RMSE. \add{Overall, synthetic benchmarks are appealing for their robustness to variation in model consistency, while consistency-based benchmarks are strong in domains where accuracy and consistency are correlated. In this setting, self-consistency is aligned with model accuracy (Figure~\ref{fig:iteration_consistency}), and per-model Pearson correlations reported in Appendix~\ref{sec:cac} further confirm that consistency reliably indicates instance-level accuracy.}


\begin{figure}[h]
        \centering
        \includegraphics[width=\textwidth]{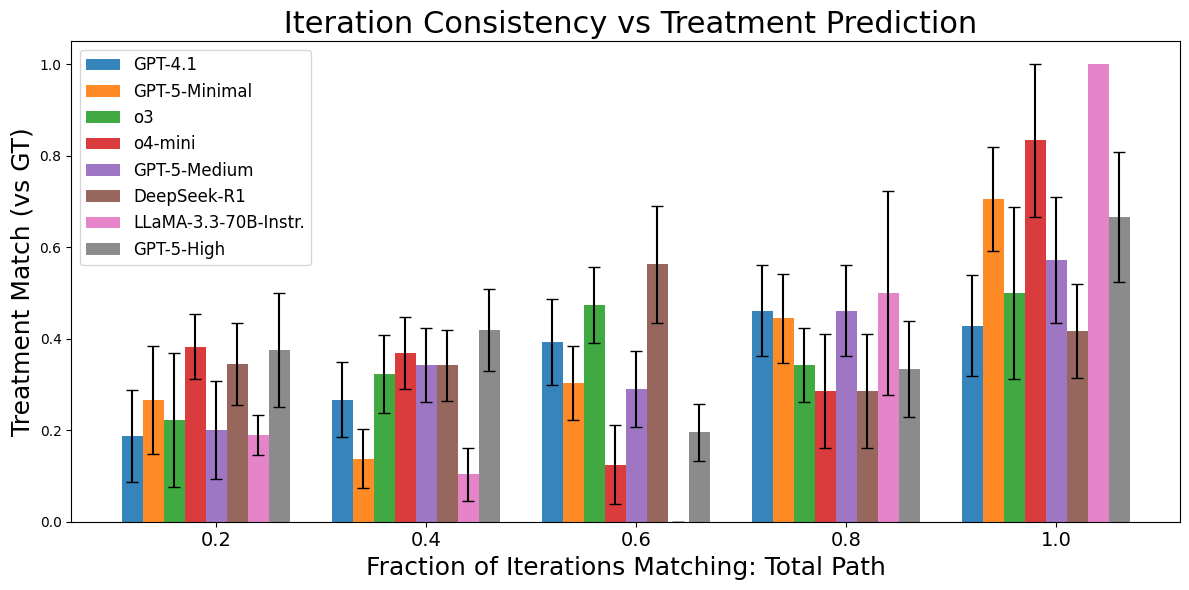}
        \label{fig:plot4}
    
    \caption{\textbf{Model accuracy increases with self-consistency across prediction runs}. Higher consistency (fraction of runs producing identical paths) correlates with improved performance for treatment matching.}
    \label{fig:iteration_consistency}
\end{figure}

\subsection{Predicting Treatment Accuracy Using Model Self-Consistency}
We develop a meta-classifier to predict when treatment recommendations are correct, using features from model consistency patterns and benchmark performance. Our approach achieves 0.804 AUROC by leveraging inference-time signals to predict accuracy without requiring ground truth labels at test time. We evaluate five feature sets to identify which signals contribute most to performance (Table~\ref{tab:feature_sets}). We stratify train and test sets \add{of the human annotated benchmark} by models and patient IDs with a 70/30 split. Performance generalizes across models and patients, enabling downstream classification with minimal human supervision.

\begin{table}[h]
\centering
\caption{Feature sets for predicting treatment recommendation accuracy}
\scriptsize
\setlength{\tabcolsep}{3pt}
\begin{threeparttable}
\begin{tabular}{lcccc}
\toprule
\makecell{\textbf{Feature} \\ \textbf{Set}} & 
\makecell{\textbf{Self-} \\ \textbf{Consistency$^*$}} & 
\makecell{\textbf{Synth.} \\ \textbf{Bench}} & 
\makecell{\textbf{Cons.} \\ \textbf{Bench}} & 
\makecell{\textbf{Cross-} \\ \textbf{Consistency$^*$}} \\
\midrule
Base       & \checkmark &             &             &               \\
Internal   & \checkmark & \checkmark & \checkmark &              \\
Agg.       &             &             &             & \checkmark \\
Base+Agg.  & \checkmark &             &             & \checkmark \\
All        & \checkmark & \checkmark & \checkmark & \checkmark \\
\bottomrule
\end{tabular}
\begin{tablenotes}
\item Agg. = Aggregated.
\item $^*$Self-consistency and Cross-Consistency features are computed per sample.
\item Synth. Bench = Synthetic Benchmarks; Cons. Bench = Consistency Benchmarks.
\end{tablenotes}
\end{threeparttable}
\label{tab:feature_sets}
\end{table}

\begin{figure}[h]
\floatconts
  {fig:roc_comparison} 
  {\caption{\textbf{Using signals from self- and cross-model consistency provides high AUC for accuracy classification.} Proxy benchmark results do not provide significant prediction signal as compared to consistency. Model AUCs range from 0.703 to 0.981, indicating strong prediction capability. LLaMA-3.3-70B-Instruct's high AUC can be attributed to its low performance on the given task, creating an arbitrary classification problem and highlighting limitations of including lower-performing models in analyses.}}
  {%
    \subfigure[ROC curves by feature set, averaged over all models.]{
      \label{fig:roc_aggregated}%
      \includegraphics[width=.7\linewidth]{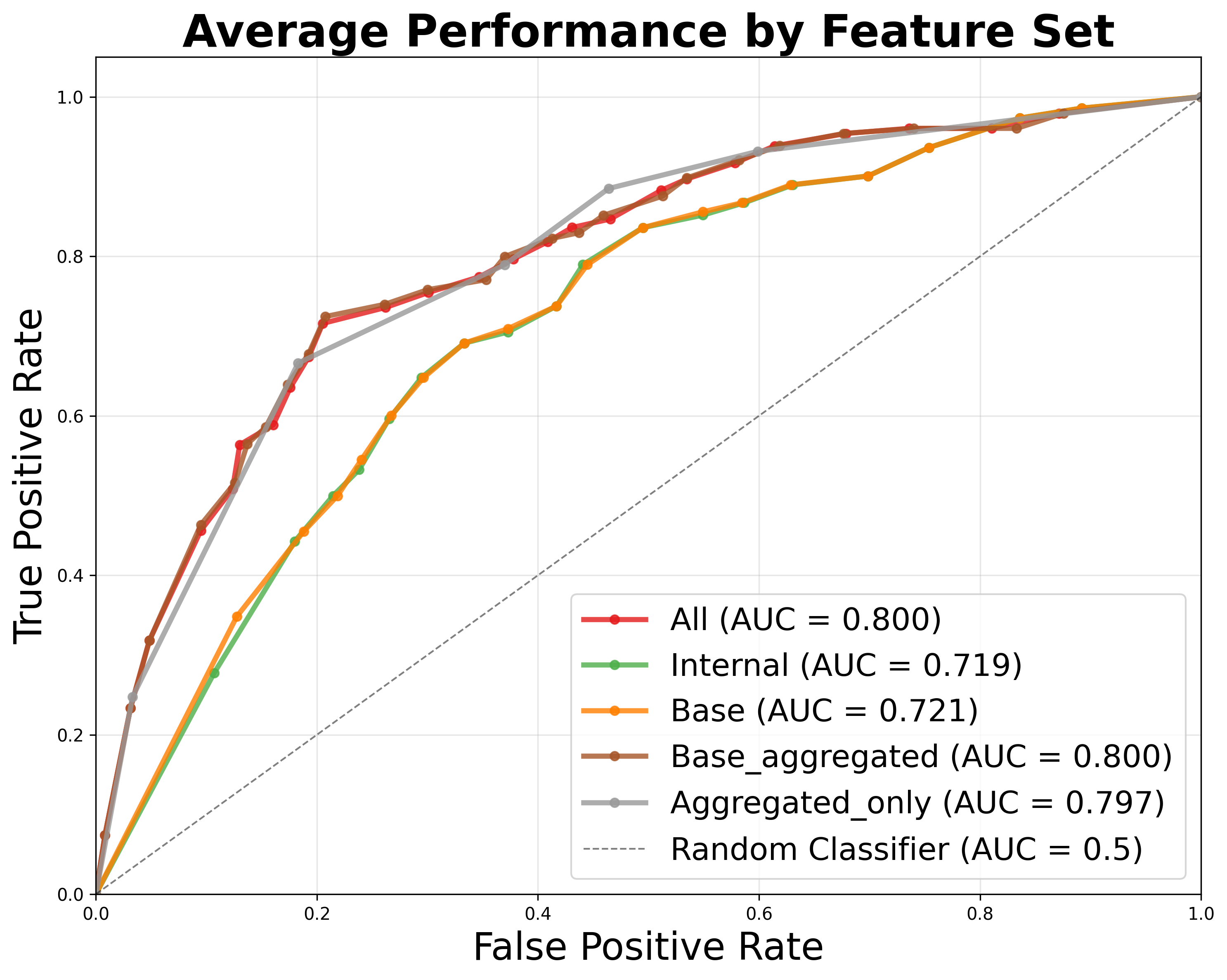}}%
    \hfill
    \subfigure[ROC curves across models for the \texttt{Base\_aggregated} feature set.]{
    \label{fig:roc_base_agg_models}%
      \includegraphics[width=.7\linewidth]{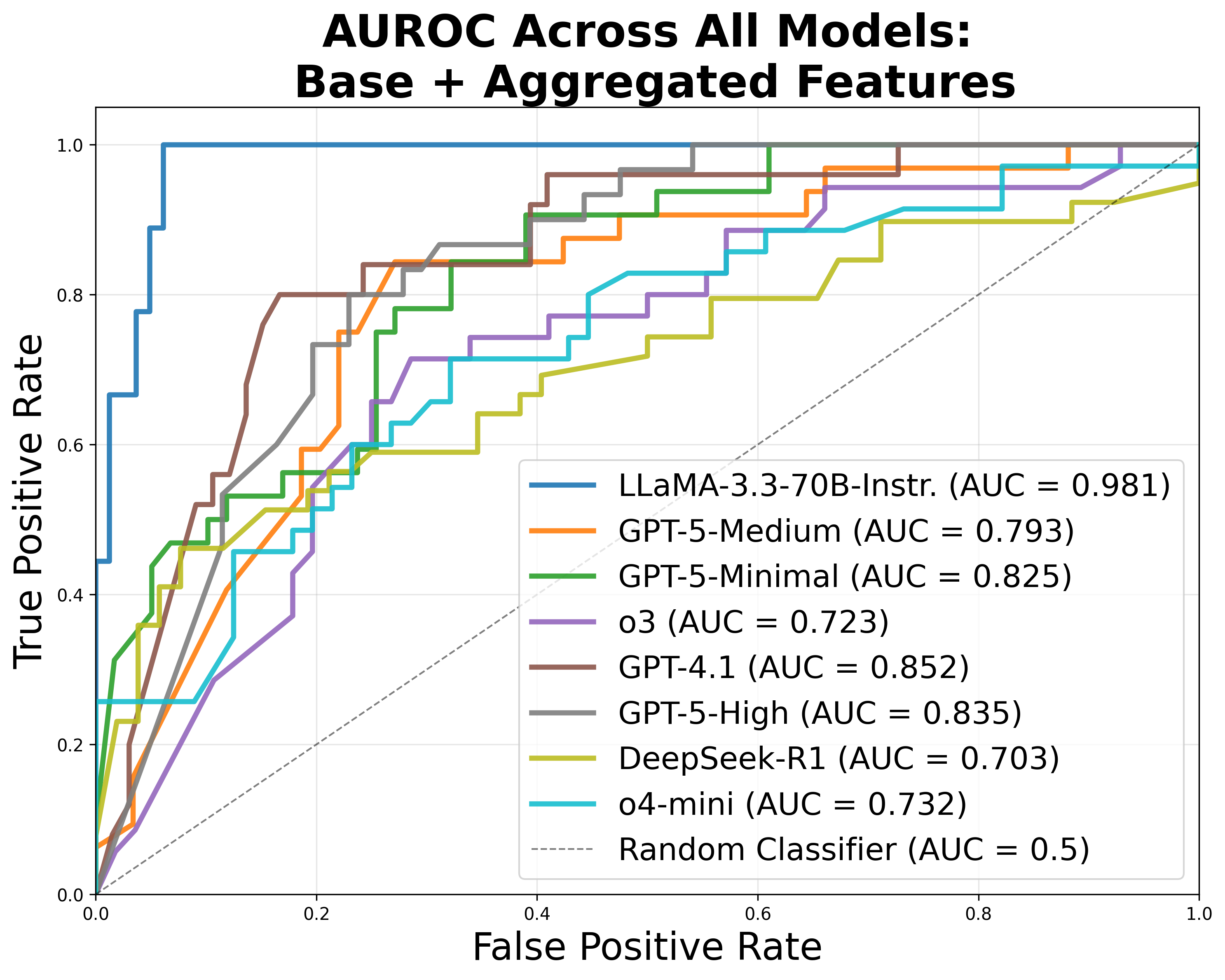}}%
    }

\end{figure}

We find in Figure~\ref{fig:roc_aggregated} that \textbf{Base\_aggregated}, \textbf{Aggregated\_only}, and \textbf{All} outperform other feature sets, indicating that cross-model consistency provides the strongest signal for accuracy classification. We see minimal gain from proxy benchmark performance, indicating consistency-based approaches are better suited for real-time analysis of model outputs than synthetic data. Even in settings without cross-model information (\textbf{Base} and \textbf{Internal}), we see high AUROC scores, indicating model self-consistency can be a strong signal for classification in isolation. Figure~\ref{fig:roc_base_agg_models} shows intra-model variability of the classifier trained using \textbf{Base\_aggregated} features, with results using \textbf{Internal} features reported in Appendix~\ref{sec:internal} \add{to highlight that even in cases where consistency and accuracy are not strongly correlated, such as for DeepSeek-R1, these features still provide some signal for accuracy classification with an AUROC of 0.582}. The strong \add{classification} performance of LLaMA-3.3-70B-Instruct \add{outputs} highlights a key issue with including lower-quality models: its performance on the NCCN task is lower than other models, arbitrarily inflating AUROC as the confidence prediction task of the meta-classifier becomes easier. Besides LLaMA-3.3-70B-Instruct, GPT-4.1 has the highest AUROC and Deepseek-R1 the lowest, indicating GPT-4.1's consistency better correlates with accuracy than Deepseek-R1.

\add{Furthermore, in a fully unsupervised setting, we are able to separate accurate from inaccurate predictions with an F1 score of 0.666 compared to the average F1 score of 0.702 in a supervised setting, demonstrating that consistency-based signals carry meaningful information for error detection without any labeled data (F1 scores for supervised classification are thresholded with value 0.5). Notably, 40.42\% of all model errors on human-labeled data can be detected without relying on any human labels (detailed analysis in Appendix~\ref{sec:error_analysis}). These results indicate that consistency signals alone provide sufficient structure to distinguish between accurate and inaccurate predictions and to identify specific failure modes, enabling model developers to detect and address potential issues prior to deployment and support iterative refinement of clinical decision support systems.}

\section{Discussion} As LLMs become increasingly capable across tasks, understanding their limitations and quantifying accuracy for specific applications is critical for deployment \citep{realworld}. This work makes three main contributions in the high-stakes medical domain. First, we formalize guideline-concordant treatment recommendation generation and introduce a dataset of 121 patient cases annotated by board-certified oncologists. Second, we compare synthetic supervision and consistency-based supervision for benchmarking models on tasks lacking human annotations, finding that consistency-based benchmarks best approximate human judgments (Spearman $r=0.88$, RMSE=0.08). Third, we develop a hybrid agent and meta-classifier trained using unsupervised features, achieving an average of 0.804 AUC on held-out models and producing confidence scores that support ROC analysis, error trade-off optimization, and regulatory compliance \citep{fda}. \add{ROC curves allow end-users—clinicians or developers—to adjust thresholds to balance sensitivity and specificity according to context, aligning with FDA guidance recommending reporting of ROC curves and related metrics, since the relative costs of false positives and false negatives vary by diagnostic scenario. Importantly, calibrated confidence scores derived from our meta-classifier provide clinicians with interpretable indicators of reliability, enabling informed decisions about the weight to assign to generated recommendations in high-stakes settings. For example, imagine an oncologist needing to recommend a second-line therapy for a patient with NSCLC. A calibrated confidence score allows the clinician to assess the reliability of the model’s guideline-adherent recommendation, which can serve as a baseline option. The clinician can then decide whether to proceed with this recommendation or intentionally deviate toward a personalized, non-guideline based approach—now with an informed understanding of the default, evidence-based standard of care. Our analysis also demonstrates that consistency signals can reveal model failure modes—such as difficulty detecting guideline non-compliance or TNM staging errors—even without extensive human annotation, highlighting both opportunities for targeted model improvement and the practical utility of these metrics in clinical contexts. Feature analysis shows that these consistency signals (self-consistency and cross-model consistency) are additionally the strongest predictors of model recommendation accuracy. This supports that consistency signals can be utilized both for identifying errors in a label-free setting and also for calibrating confidence to allow users to make informed decisions regarding their trust in the system output.}

Several promising directions emerge from this work. Expanding the dataset to other cancer types and guidelines would help characterize the generalizability of our consistency-based benchmarking approach, potentially enabling cross-domain accuracy prediction \citep{cancer1, cancer2}. Further evaluation of self- and cross-model consistency as a signal for accuracy classification across broader model sizes and architectures could deepen understanding of this method’s robustness. Increasing both the size of the human-annotated dataset and the number of dually annotated samples would strengthen confidence in ground-truth labels. Since clinicians exhibit variability in path selection, future work should also explore ways to explicitly model and incorporate human uncertainty into evaluation. Adaptive learning approaches that account for uncertainty across clinical scenarios may further improve sensitivity of proxy metrics \citep{active1,active2}. Finally, investigating how proxy benchmark data can be leveraged for alignment with downstream human preferences could help mitigate data bottlenecks that limit current methods \citep{proxy1,proxy2}.

\bibliography{jmlr-sample}

\appendix
\section{Annotation Details}
\label{subsec:annotation}
We begin by extracting the decision tree pages from the NCCN NSCLC guidelines \citep{NCCN} and creating a JSON-based representation where each decision point is labeled as a ``node." Terminal nodes with ``recommendation" labels contain final treatment recommendations for clinicians, while intermediate nodes represent decision points requiring additional clinical information.

To evaluate model capability in mapping these guidelines onto real clinical data, we created a human-annotated dataset for guideline-adherent reasoning in NSCLC. Expert annotators traced patient notes through the NCCN guideline decision tree, recording the ordered sequence of page–node identifiers (e.g., NSCL-1-1 → NSCL-2-1 → … → terminal). Each annotation aimed to reach either the appropriate ``recommendation'' node when sufficient clinical information was available, or the furthest node that could be accurately determined given the available patient data. Annotators additionally marked whether the observed care was guideline-compliant and provided brief rationales for any identified as non-compliance. \add{Non-compliance is treated as a valid ``path'' and evaluates model's capacity to identify patient treatment trajectories as non-compliant with given guidelines.}

Thirteen physicians performed the annotations: twelve board-certified oncologists and one hematology–oncology fellow, averaging 13 years of clinical experience. Participants were recruited through data vendors, screened for oncology knowledge and rubric literacy, and verified via official certification registries. All physicians were compensated, and annotation instructions are provided.

A subset of cases was double-annotated, and disagreements were resolved through adjudication meetings, with all decisions documented for consistency. A strict no-AI-use policy was enforced via Insightful time-tracking with periodic screenshot audits \citep{insightful}. Annotation time averaged 46.5 minutes per case. All cases were fully de-identified and contained no protected health information, so the study did not constitute human-subjects research and no IRB was required. 
\newpage
\begin{figure*}[htbp]
\centering
\begin{tcolorbox}[
    colback=gray!10,
    colframe=gray!50,
    title={\textbf{Clinical Annotation Instructions}},
    fonttitle=\bfseries,
    boxrule=1pt,
    rounded corners,
]

You are helping us understand how a synthetic patient progressed through the NCCN guidelines for non-small cell lung cancer (NSCLC), based on information in their clinical note.

Your task is to carefully read the clinical note and determine which steps in the NCCN guideline the patient appears to have gone through. Using your clinical judgment and the structure of the guideline, trace the path the patient has followed so far.

\textbf{Path Format:}
You will identify the treatment path in the following format:
\begin{center}
NSCL-x-y $\rightarrow$ NSCL-a-b $\rightarrow$ \ldots $\rightarrow$ Final Treatment Node
\end{center}

Each step should include the full node ID from the guideline (e.g., NSCL-1-1). Do not skip any nodes, even if some are broader or less specific than others. The final step should be a terminal node—a node that includes a clear treatment recommendation.

If the patient's treatment has not reached a terminal node, you should provide the recommended pathway to the next terminal node, as permitted by the information in the note.

\textbf{Compliance Assessment:}
There is a box to check if the patient's path complies with the NCCN guidelines. If you believe the patient's path does not align with the NCCN guideline, please do not check the box. In this case, you will also include a brief explanation of why the treatment path doesn't match the guidelines. Only include a ``reason'' if the path is not compliant. If the path is guideline-compliant, just provide the full treatment path.

\textbf{Important Notes:}
\begin{itemize}
    \item You do not need to speculate beyond what is described in the note
    \item You do not need to assume the patient has already received the final treatment—just determine the most appropriate next step or current position in the guideline  
    \item In some cases, the NSCLC Guidelines may not apply. For these cases, please note that the guidelines do not apply and proceed to the next task
\end{itemize}

\textbf{Materials Provided:}
\begin{itemize}
    \item A clinical note (the patient's case)
    \item The full NCCN guideline (for reference)
\end{itemize}

\textbf{Tips and Tricks:}
We recommend utilizing ``Command + f'' (Mac) or ``Ctrl+f'' (Windows) to find key terms and parts of the note/guideline tree that are relevant. The guidelines consist of many ``pages'' (i.e. NSCLC-1, NSCLC-2) which contain further questions that will bring you to other pages.

\textbf{Example Workflow:}




\textit{Step 1:} \textbf{REDACTED DUE TO NCCN LICENSING REGULATIONS}

\textit{Step 2:} Using Ctrl+F (Command+F on Mac), proceed to the next step based on the patient's clinical notes.

Per the example above, search ``NSCL-2'' to skip to the section beginning: \texttt{``NSCL-2'': \{ ``page\_id'': ``NSCL-2''}

Under NSCL-2, you see ``nodes'' and then a numbered list of different criteria. Begin with node 1 and follow the branching guidance from there.

At each step, confirm that the treatment outlined in the patient's clinical note is in accordance with the NCCN Guidelines. If the treatment diverges from the path outlined by the guidelines, it would be considered noncompliant. The treatment path you provide should be the treatment path that was followed in the patient's care. If this differs from the path recommended by the NCCN Guidelines, indicate that the path was noncompliant.

\end{tcolorbox}
\label{fig:annotation-instructions}
\end{figure*}

\section{Path Overlap Score}
\label{subsec:path_overlap}

Let $\mathcal{P} = \{P_1, P_2, \ldots, P_k\}$ be a collection of $k$ predicted paths, where 
\[
P_i = (p_{i,1}, p_{i,2}, \ldots, p_{i,n_i}), \quad i = 1, \ldots, k,
\]
and $p_{i,j}$ denotes the $j$-th node in path $P_i$.  

Define the node set of path $P_i$ as
\[
V_i = \{p_{i,1}, p_{i,2}, \ldots, p_{i,n_i}\},
\]
and the union and intersection of all paths as
\[
U = \bigcup_{i=1}^k V_i, \quad I = \bigcap_{i=1}^k V_i.
\]

The \textbf{Path Overlap Score} (corresponding to \texttt{path\_match\_fraction} in code) is defined as the Jaccard similarity:
\[
\text{Overlap}(\mathcal{P}) = 
\begin{cases}
\frac{|I|}{|U|}, & |U| > 0, \\
1, & |U| = 0.
\end{cases}
\]

\subsection*{Properties}
\begin{itemize}
    \item $\text{Overlap}(\mathcal{P}) \in [0,1]$.
    \item $\text{Overlap}(\mathcal{P}) = 1$ if and only if all paths share exactly the same set of nodes.
    \item Higher scores indicate greater similarity in node coverage across paths.
    \item Symmetric with respect to path ordering.
    \item Sensitive only to node membership, not ordering or repetition.
\end{itemize}

\section{Treatment Match}
\label{subsec:treatment_match}

\section*{Final Treatment Consistency Score}

Let $\mathcal{P} = \{P_1, P_2, \ldots, P_k\}$ be a set of $k$ predicted treatment paths for a given patient, where each path $P_i = (p_{i,1}, p_{i,2}, \ldots, p_{i,n_i})$ terminates with a final treatment recommendation $f_i = p_{i,n_i}$.

\subsection*{Case 1: Ground Truth Available}

When ground truth final treatment $f^*$ is available, the score is computed as the indicator function:
$$\mathbf{1}_{f_i = f^*} = \begin{cases} 
1 & \text{if } f_i = f^* \\
0 & \text{otherwise}
\end{cases}$$

This returns whether there was an exact match between the ground truth and the prediction final treatment.

\subsection*{Case 2: No Ground Truth Available}

In the absence of ground truth, we measure internal consistency by computing the proportion of repeated final treatments:

$$S_{\text{final}}(\mathcal{P}) \;=\; \frac{\max_{t \in T} c(t)}{k}$$

where:
$$k = \text{total rollouts}$$
$$c(t) = \sum_{i=1}^{k} \mathbf{1}_{\{f_i = t\}}$$

The numerator $\max_{t \in T} c(t)$ counts the total number of ``repeated" final treatment occurrences (i.e., how many times each treatment appears beyond its first occurrence) and returns the frequency of the mode treatment selection. For all models $k=10$.

\subsection*{Properties}

\begin{itemize}
\item $S_{\text{final}}(\mathcal{P}) \in [0, 1]$ for both cases
\item Higher scores indicate better accuracy (Case 1) or higher consensus (Case 2)
\item \textbf{Case 2}: Score of 0 indicates complete disagreement; score approaching 1 indicates strong consensus
\end{itemize}

\section{Error Analysis}
\label{sec:error_analysis}
We find that 40.42\% of all errors made by models on human-annotated clinical text are flagged within the top 5 points of confusion by the model itself during its internal rollouts.  This enables practitioners to refine instructions, perform fine-tuning, or apply other interventions to address these frequent errors—without relying on human annotations to detect them. Figure~\ref{fig:ea_comparison} shows the overlap between errors identified by model consistency vs. the true errors made by the model against human annotated data. Discrepancies mostly arise around guideline compliance and tumor staging, indicating that models struggle to differentiate between non-compliant and compliant cases with high accuracy as well as are unable to leverage parametric knowledge to perform tumor staging. Potential interventions could be training a model specifically for either task or providing additional context with clear instructions regarding clinical expectations.

\begin{figure*}[!htbp]
\floatconts
  {fig:ea_comparison} 
  {\caption{Identification of most common discrepancies between human annotations and model annotations compared to most common discrepancies between $k$ model rollouts of path prediction.}}
  {%
    \subfigure[GPT-5 Medium]{%
      \label{fig:1a}%
      \includegraphics[width=0.45\linewidth]{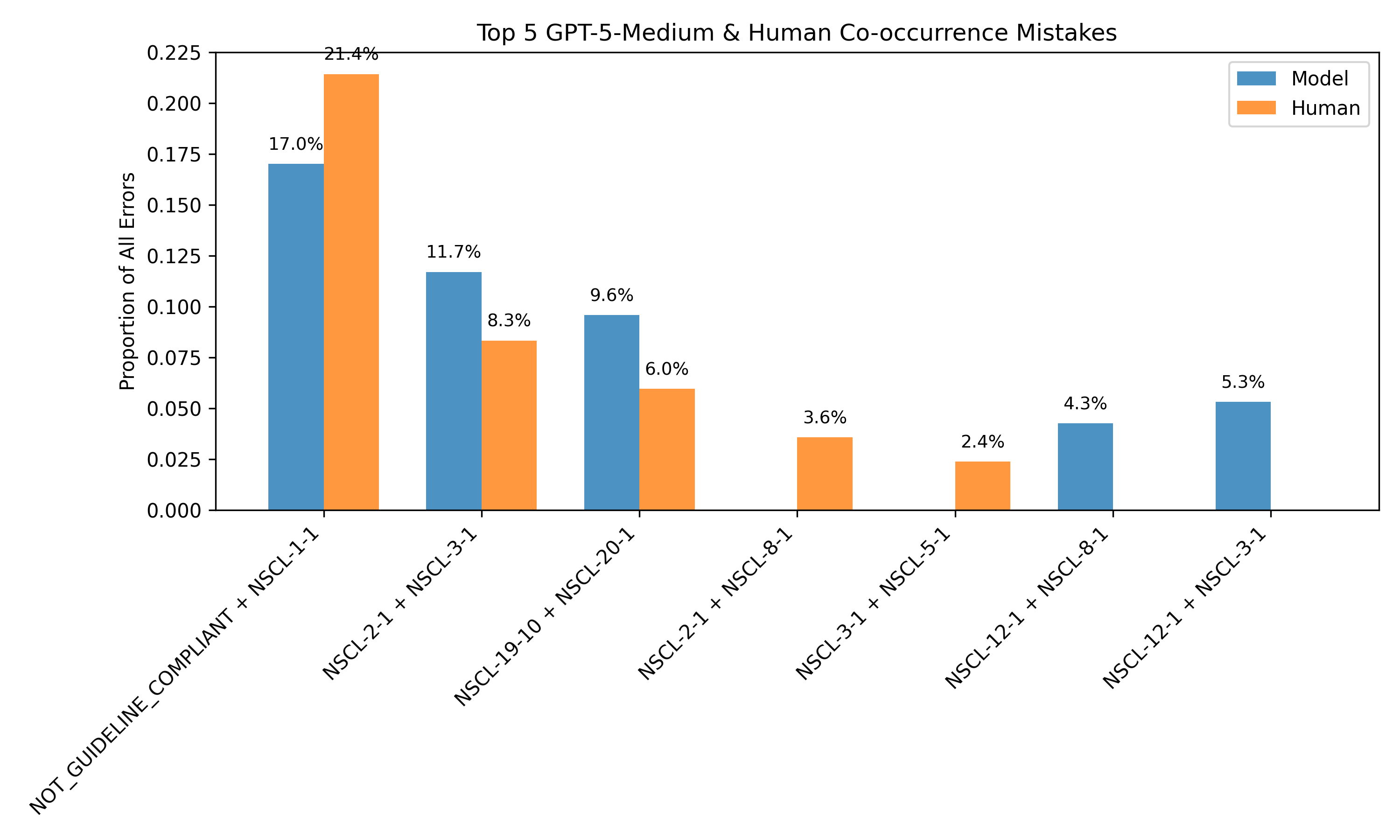}}%
    \hfill
    \subfigure[GPT-5 Minimal]{%
      \label{fig:1b}%
      \includegraphics[width=0.45\linewidth]{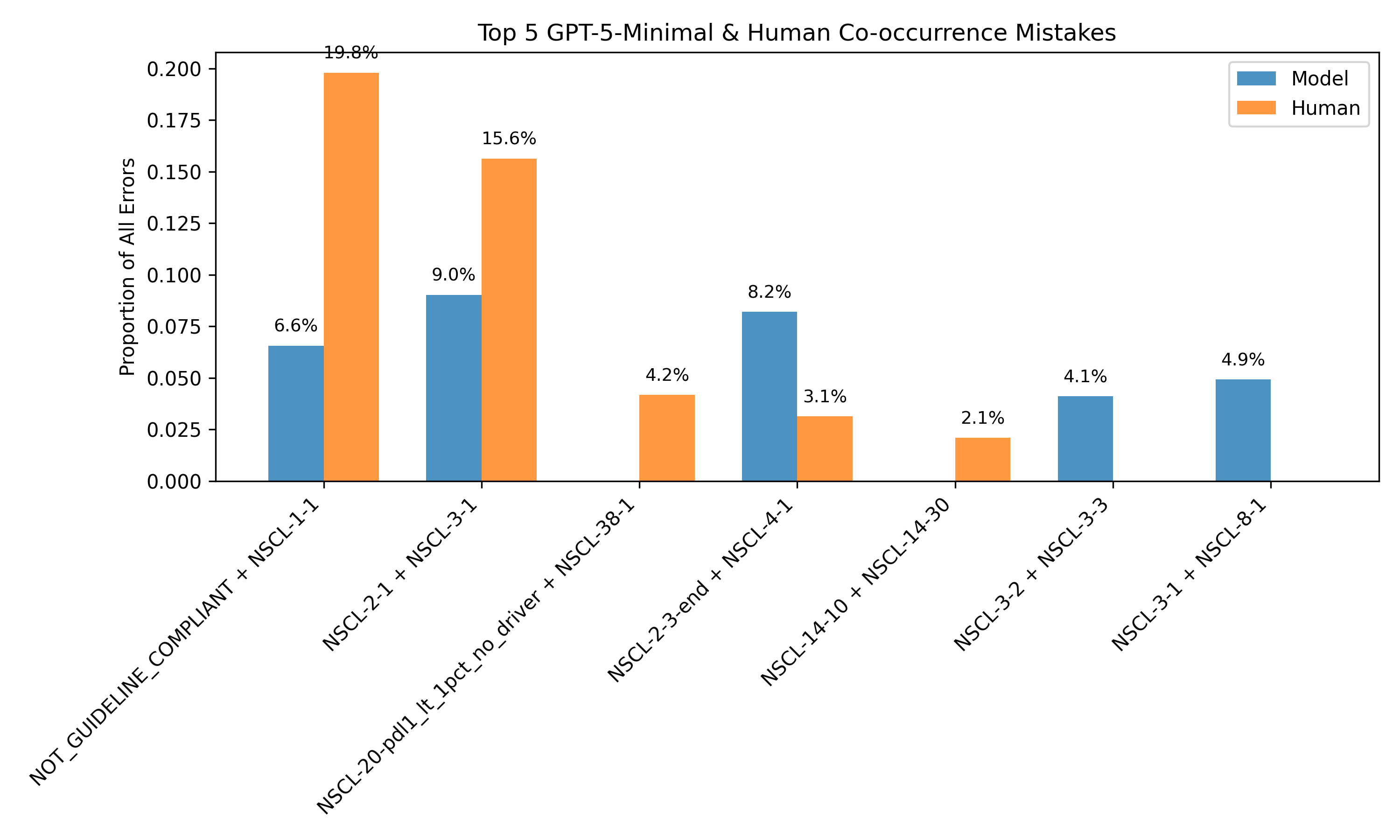}}%

    \subfigure[GPT-4.1]{%
      \label{fig:1c}%
      \includegraphics[width=0.45\linewidth]{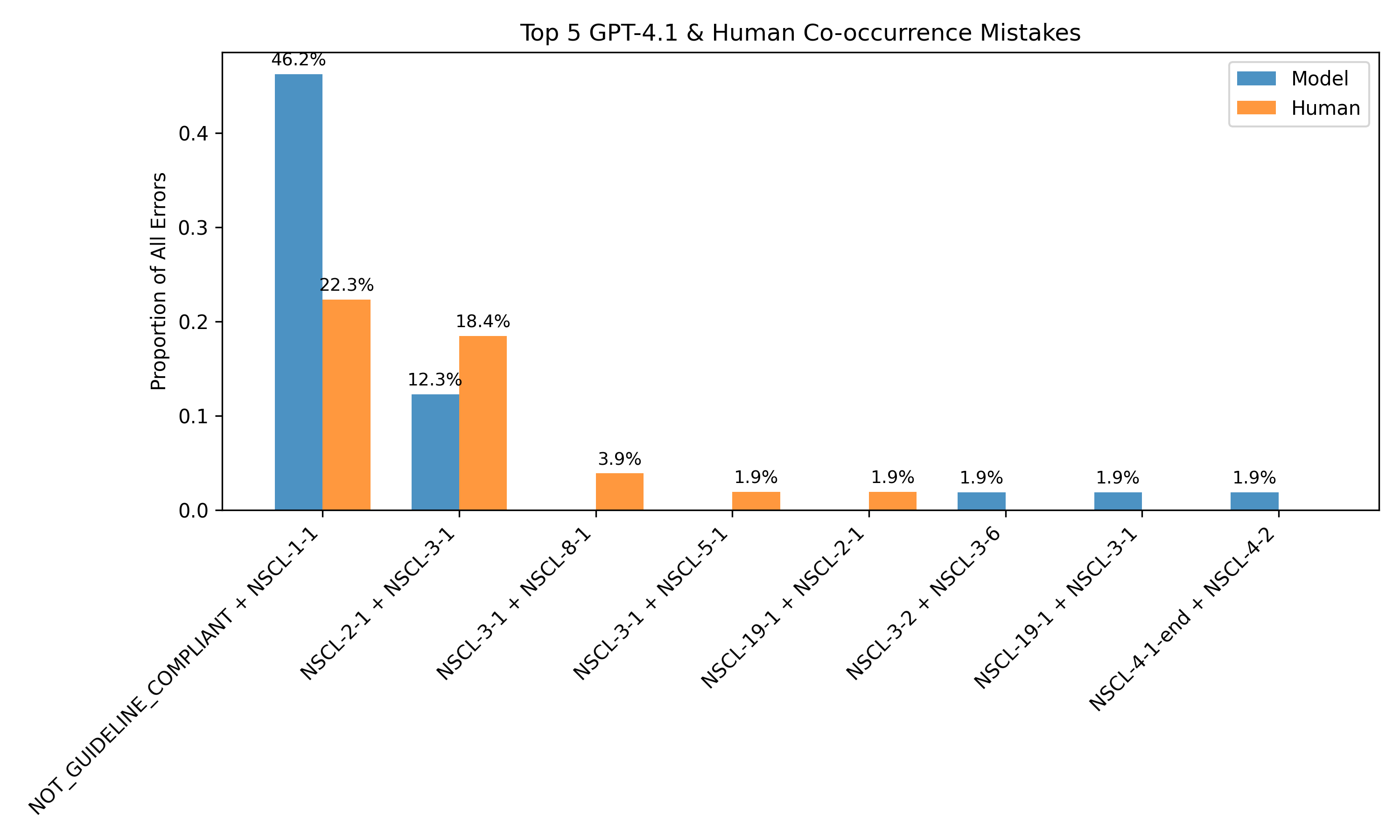}}%
    \hfill
    \subfigure[o3]{%
      \label{fig:2a}%
      \includegraphics[width=0.45\linewidth]{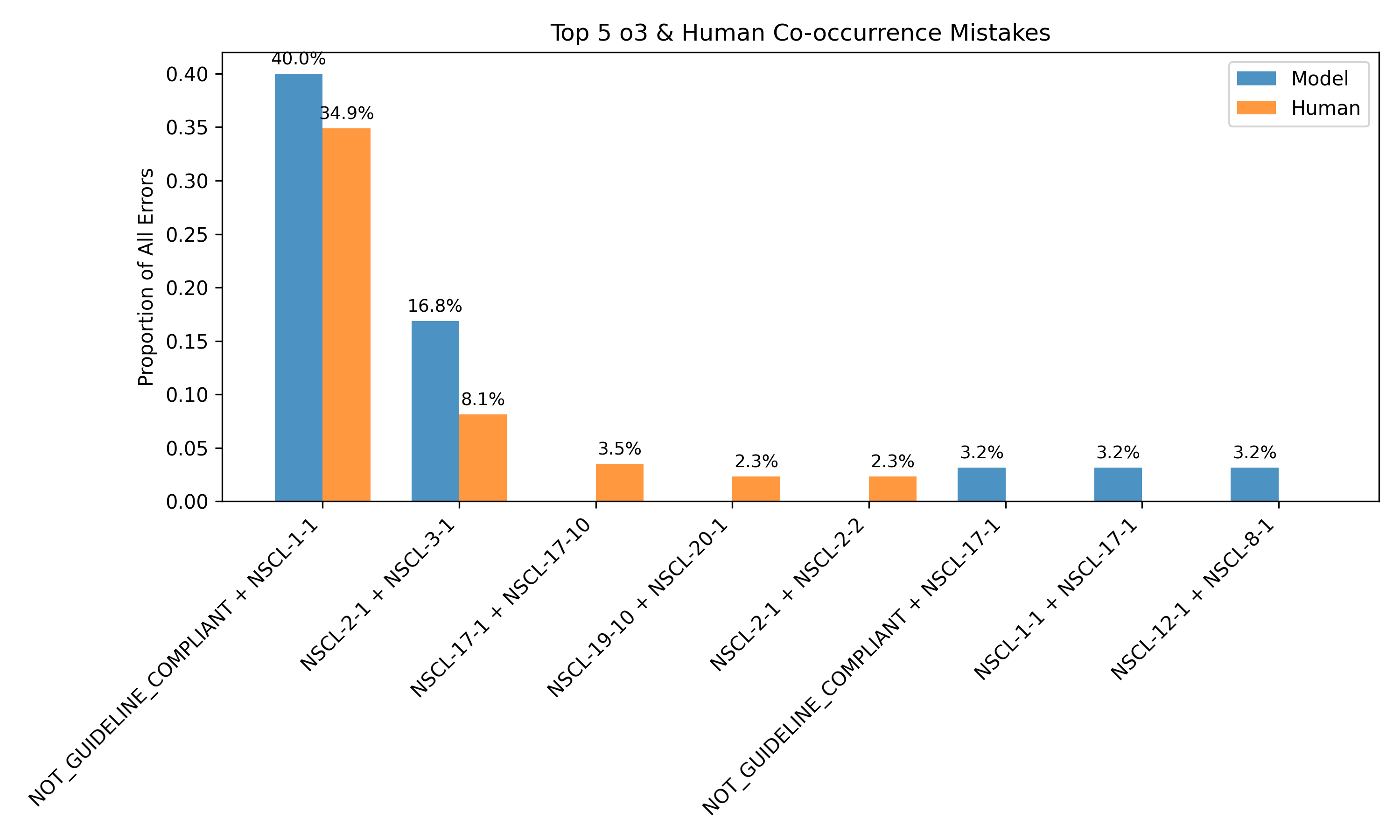}}%

    \subfigure[o4-mini]{%
      \label{fig:2b}%
      \includegraphics[width=0.45\linewidth]{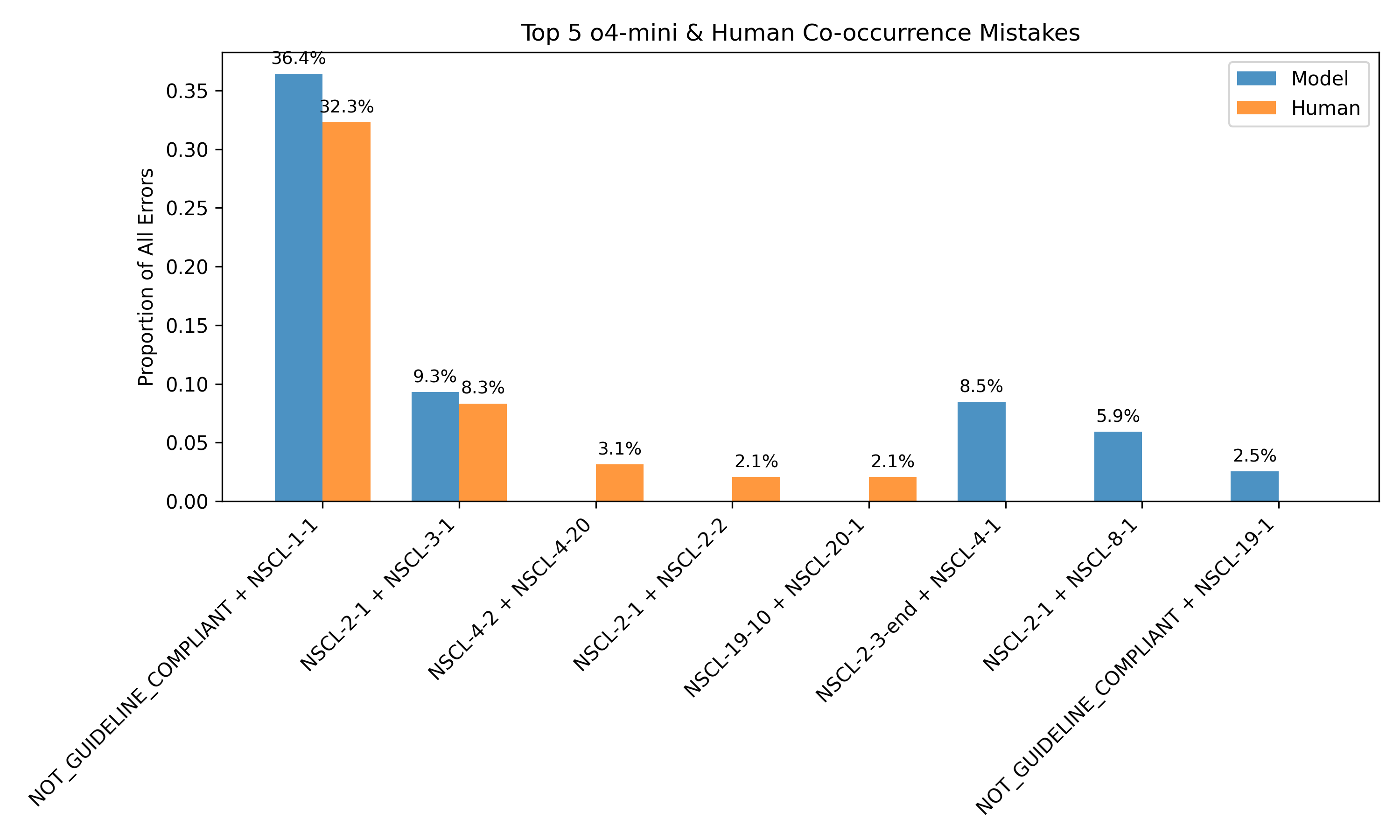}}%
    \hfill
    \subfigure[DeepSeek-R1]{%
      \label{fig:2c}%
      \includegraphics[width=0.45\linewidth]{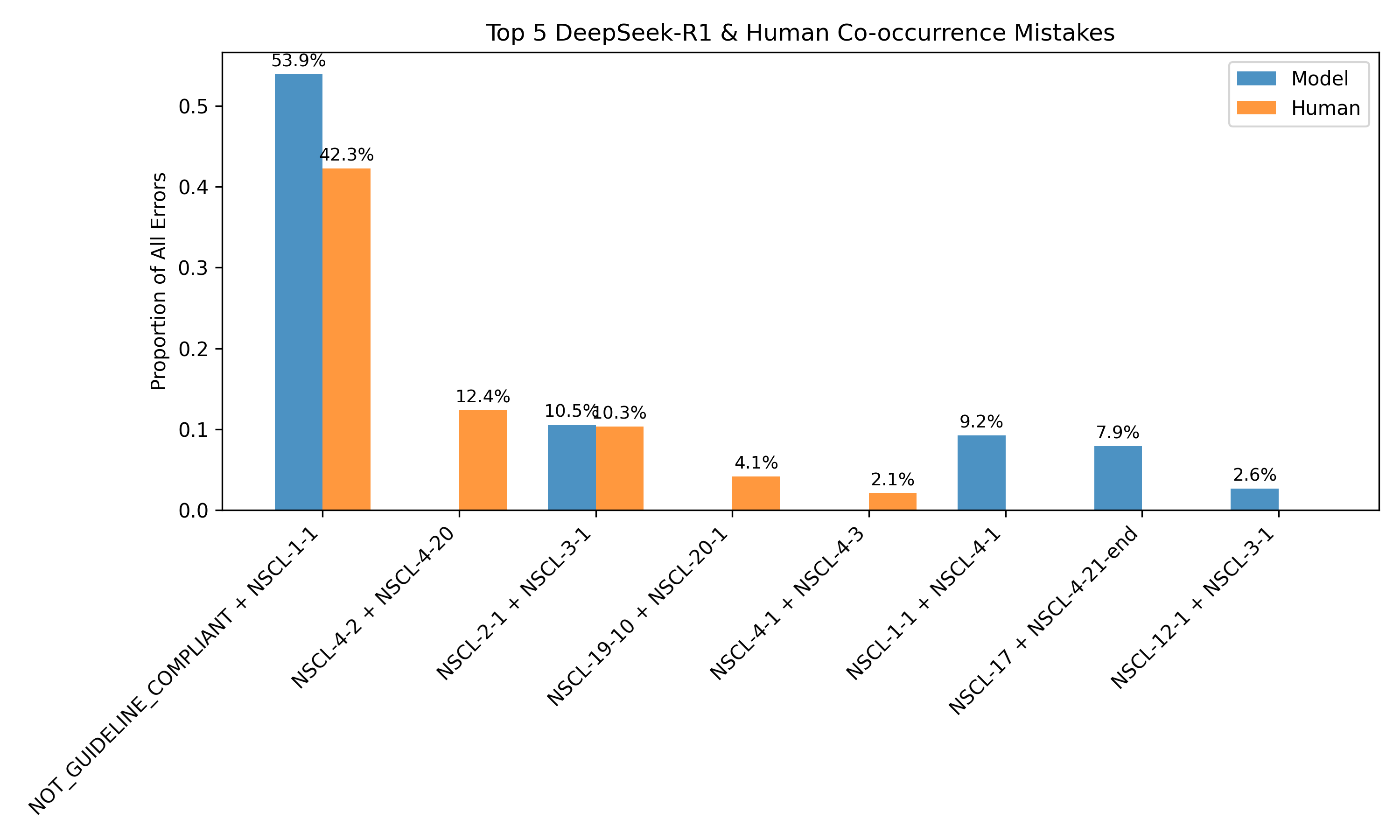}}%

    \subfigure[LLaMA-3.3-70B-Instruct]{%
      \label{fig:2d}%
      \includegraphics[width=0.45\linewidth]{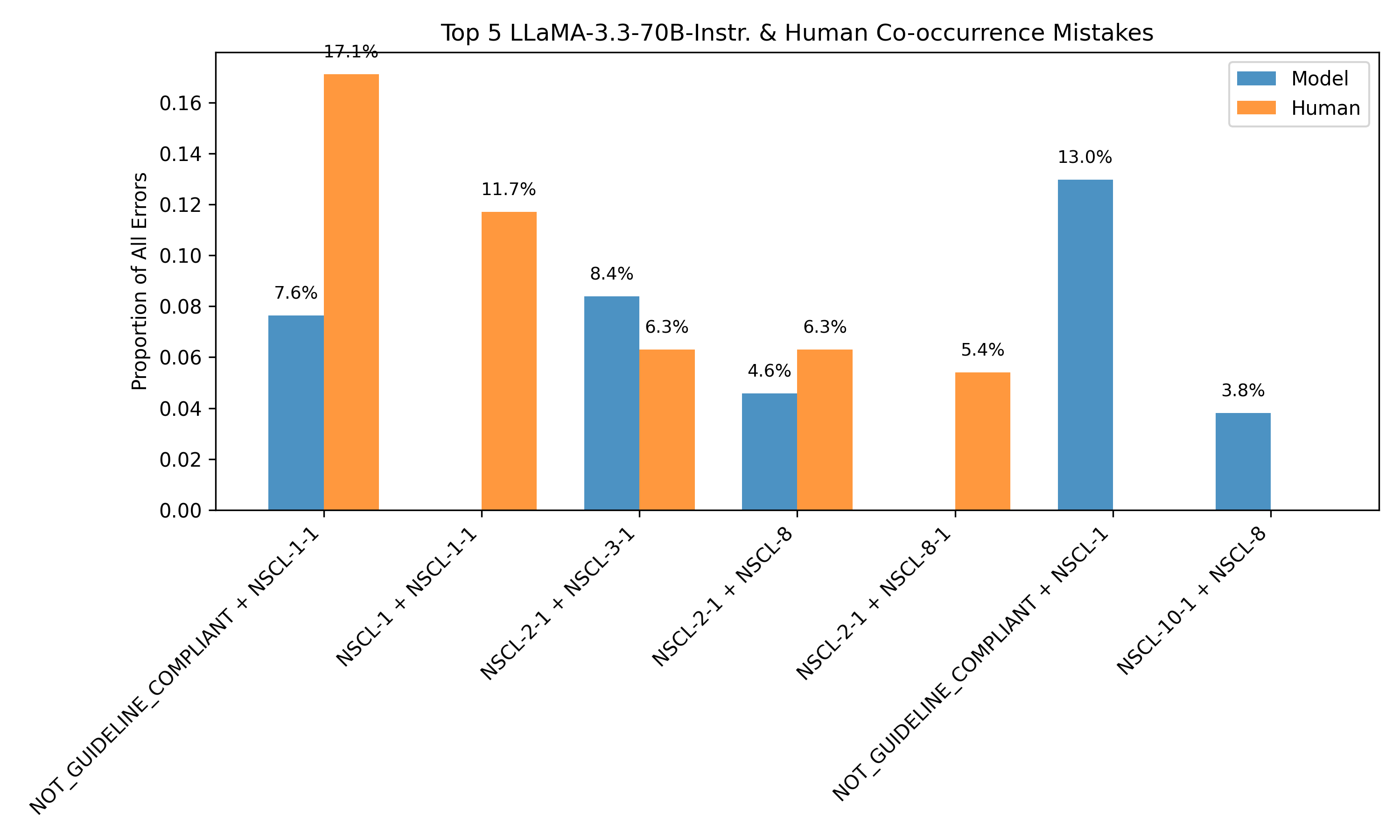}}%
    \hfill
    \subfigure[GPT-5 High]{%
      \label{fig:2e}%
      \includegraphics[width=0.45\linewidth]{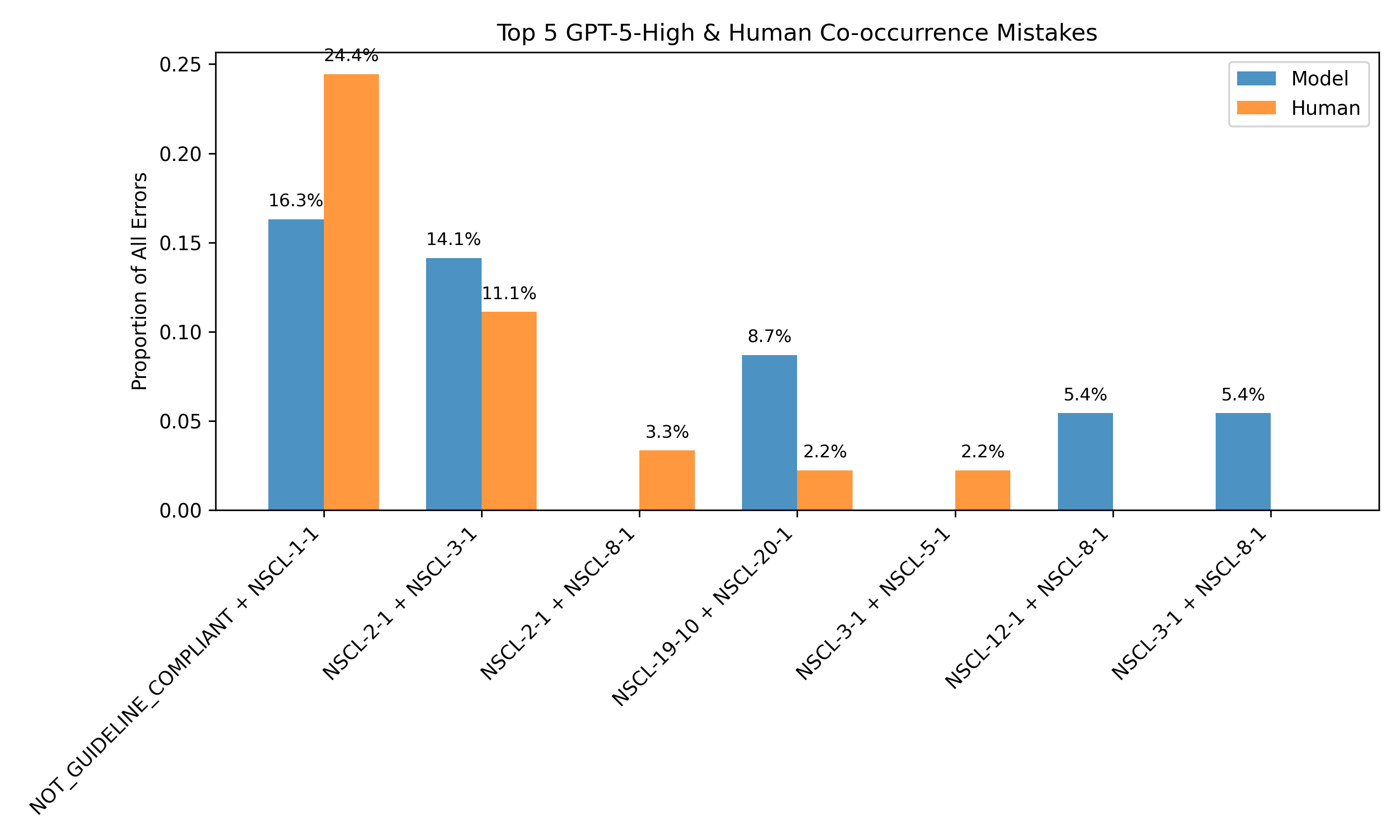}}%
  }
\end{figure*}

\section{Consistency-Accuracy Correlation}
\label{sec:cac}
We show that for both Path Overlap metric and Treatment Match metrics, consistency of a prediction is correlated with accuracy (Table~\ref{tab:r_scores_consistency}). We do see, however, that there are different levels of correlation, indicating that consistency-based approaches inherently are more accurate for some model families than others. We note that DeepSeek-R1 has an outlier correlation coefficient value, being negative for Path Overlap score and extremely low for Treatment Match score's correlation with accuracy. This highlights the robustness of the meta-classification pipeline to varying model internal consistencies. Despite the lack of direct correlation between self-consistency and accuracy for this model, we are still able to classify accuracy with high fidelity.

\begin{table}[!htbp]
\centering
\caption{Iteration consistency vs. accuracy correlation ($r$) for path overlap score and treatment match score. Mean correlation values across all models $\pm$ SEM are reported at the bottom.}
\scriptsize
\begin{tabular}{lcc}
\hline
\makecell{\textbf{Model}} & 
\makecell{\textbf{Path} \\ \textbf{Overlap}} & 
\makecell{\textbf{Treatment} \\ \textbf{Match}} \\
\hline
GPT-5-High     & 0.477 & 0.472 \\
GPT-4.1        & 0.812 & 0.923 \\
GPT-5-Minimal  & 0.925 & 0.866 \\
o3             & 0.700 & 0.795 \\
o4-mini        & 0.566 & 0.491 \\
GPT-5-Medium   & 0.794 & 0.935 \\
DeepSeek-R1    & -0.647 & 0.129 \\
LLaMA-3.3-70B-Instr.  & 0.688 & 0.789 \\
\hline
\textbf{Mean $\pm$ SEM} & 0.540 $\pm$ 0.179 & 0.675 $\pm$ 0.103 \\
\hline
\end{tabular}
\label{tab:r_scores_consistency}
\end{table}

\section{Unsupervised clustering performance}
\begin{figure}[h]
    \centering
    \includegraphics[width=\textwidth]{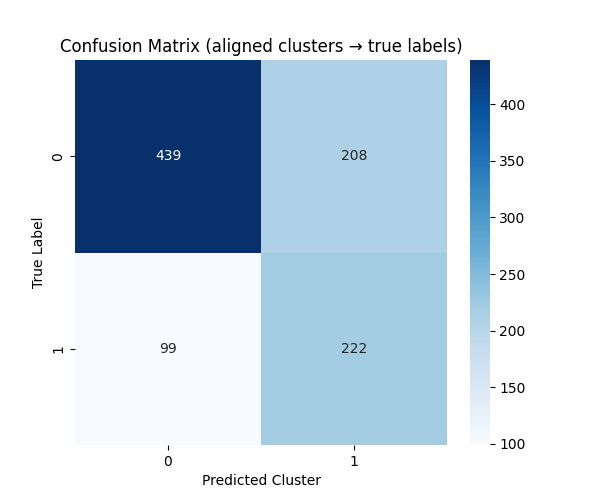} 
    \caption{Confusion Matrix for K-means clustering over consistency-derived features.}
    \label{fig:unsupervised}
\end{figure}

We cluster on features related to self- and cross-model consistency (Figure~\ref{fig:unsupervised}). We are able to distinguish between True Negatives and False Negatives with high accuracy, and are able to derive some signal regarding the True Positives. This indicates that consistency alone can be used as a baseline in an unsupervised method to identify accuracy of prediction. The performance is below supervised approaches, as to be expected, but is promising for further applications in which supervision is not viable or possible. F1 score is .666, indicating potential of unsupervised methods in conjunction with consistency approaches to evaluate model performance in zero-label settings.

\section{Meta-Classifier Performance on Internal Features}
\label{sec:internal}
\begin{figure}[h]
    \centering
    \includegraphics[width=\textwidth]{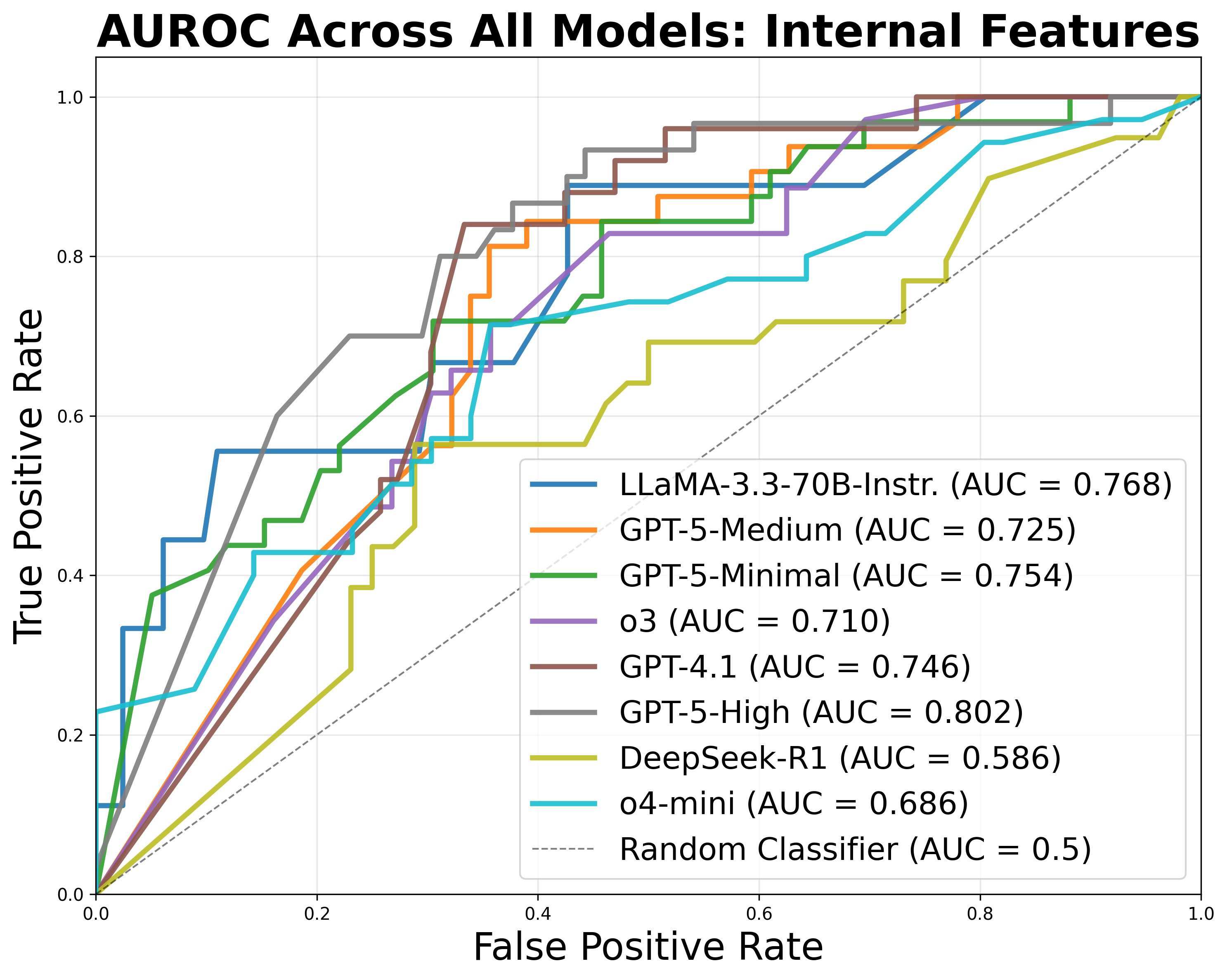} 
    \caption{ROCs disaggregated by model for classification of treatment prediction accuracy using exclusively internal features during training.}
    \label{fig:unsupervised_appendix}
\end{figure}

We report the performance of our meta-classifier using internal features only for training, as shown in Figure~\ref{fig:unsupervised_appendix}. This enables us to disaggregate classification performance between cross-model and self consistency. We see here that the AUROC scores for each model is lower than when we include cross-model information, but it is still significantly above random, indicating that while cross-model consistency is a useful feature, it is not the only feature that allows for insight into model performance.

\section{Model Information}
For supervised classification, we trained a logistic regression classifier with L2 regularization (C=0.1), optimized using LBFGS with up to 10,000 iterations, and applied class-balanced weighting to account for label imbalance. Table~\ref{tab:model_info} reports further API details. 

\begin{table*}[htbp]
\centering
\caption{Model release dates and API versions.}
\label{tab:model_info}
\begin{tabular}{lcc}
\hline
\textbf{Model} & \textbf{Provider} & \textbf{API Version} \\
\hline
DeepSeek-R1     & Azure AI & 2024-05-01-preview \\
LLaMA-3.3-70B-Instr.  & Azure OpenAI     & 5 (Model release: 2024-12-06) \\
GPT-4.1         & Azure OpenAI & 2024-12-01-preview \\
GPT-5           & Azure OpenAI & 2024-12-01-preview \\
o4-mini         & Azure OpenAI & 2024-12-01-preview \\
o3              & Azure OpenAI & 2024-12-01-preview \\
\hline
\end{tabular}
\end{table*}

\add{\section{Human Annotation Disagreement Analysis}}
\label{sec:disagreement_analysis}

Of the 11 dually annotated records:

\begin{enumerate}
    \item 7 were agreed upon by all physicians.
    \item 4 required adjudication
    \item 2 involved disagreements about whether the patient’s history complied with guidelines (these cases are excluded from our benchmark).
    \item 1 involved assumptions regarding the patient’s history; after clarification to rely strictly on the provided record, both clinicians agreed.
    \item 1 involved a TNM staging error by one clinician, resolved upon discussion.
\end{enumerate}

After removing non-compliant cases, agreement across the benchmark was 7/9. This reflects the inherent difficulty and nuance of the task, setting a realistic upper bound for LLM performance in clinical decision support scenarios.

\add{\section{NCCN Tree Generation}}
To construct the clinical decision tree, we first manually structured a subset of sample guideline pages into a JSON-based hierarchical format representing decision nodes and treatment branches. These manually curated examples served as few-shot demonstrations for prompting the language model to generate additional decision trees for the remaining guideline sections. The generated JSON outputs were automatically validated to ensure structural integrity and adherence to the required schema. While this process provided an initial level of validation, the resulting trees were not independently verified by clinical experts, and thus may contain minor inconsistencies or omissions relative to expert-curated guideline representations. The generated NCCN tree for NSCLC has 3,302 nodes, 3,301 edges, and a max depth of 8.

\add{\section{Clinical Data Details}}
The patient data used for analysis comprised approximately 2,900 patients diagnosed with breast and lung cancer at various stages, drawn from two U.S. health systems: a community-based system and a large nonprofit health system, each spanning multiple hospital locations. The dataset included electronic medical records, radiology reports, pathology reports, and other clinical documents such as next-generation sequencing (NGS) reports. Tumor registry data were obtained from three complementary sources: (1) the hospitals’ internal registries, (2) registry submissions to the state (for one system), and (3) manually labeled registry data extracted by over 30 registered oncology nurses. All data were de-identified by the vendor prior to delivery, following HIPAA-compliant procedures that included redaction of all PHI, date shifting relative to the date of birth, and removal of DICOM headers and pathology metadata containing identifiers.

\end{document}